\documentclass[runningheads]{llncs}
\usepackage{graphicx}

\usepackage{tikz}
\usepackage{comment}
\usepackage{amsmath,amssymb} 
\usepackage{color}
\usepackage[symbol]{footmisc}

\usepackage{booktabs}
\usepackage{adjustbox}
\usepackage{hyperref}
\hypersetup{
  colorlinks   = true, 
  urlcolor     = red, 
  linkcolor    = blue, 
  citecolor   = green 
}

\usepackage{bm}
\usepackage{pifont}
\usepackage{xcolor}
\usepackage{array}
\usepackage{microtype}
\usepackage{wrapfig}

\DeclareMathOperator*{\argmax}{argmax}
\newcommand{\cmark}{{\color{green}\ding{52}}}
\newcommand{\xmark}{{\color{red}\ding{56}}}
\newcolumntype{P}[1]{>{\centering\arraybackslash}p{#1}} 
\newcolumntype{M}[1]{>{\centering\arraybackslash}m{#1}} 

\newcommand\blfootnote[1]{%
  \begingroup
  \renewcommand\thefootnote{}\footnote{#1}%
  \addtocounter{footnote}{-1}%
  \endgroup
}

\usepackage[accsupp]{axessibility}  

\usepackage[width=122mm,left=12mm,paperwidth=146mm,height=193mm,top=12mm,paperheight=217mm]{geometry}

\begin{document}
\pagestyle{headings}
\mainmatter
\def\ECCVSubNumber{3781}  

\title{TAVA: Template-free Animatable\\Volumetric Actors} 

\titlerunning{TAVA: Template-free Animatable Volumetric Actors}
%
\author{Ruilong Li$^{1,3}$, Julian Tanke$^{2,3}$, Minh Vo$^3$, Michael Zollh\"ofer$^3$,\\J\"urgen Gall$^2$, Angjoo Kanazawa$^1$, Christoph Lassner$^3$}
\institute{\blfootnote{$^{\dagger}$Work was done partially while Ruilong and Julian were at Meta Reality Lab.}1. UC Berkeley\quad 2. University of Bonn\quad 3. Meta Reality Labs Research}
\authorrunning{Li et al.}
%
\maketitle
\vspace*{-2em}
\begin{abstract}
Coordinate-based volumetric representations have the potential to generate photo-realistic virtual avatars from images.
However, virtual avatars also need to be controllable even to a novel pose that may not have been observed. 
Traditional techniques, such as LBS, provide such a function; yet it usually requires a hand-designed body template, 3D scan data, and limited appearance models.
On the other hand, neural representation has been shown to be powerful in representing visual details, but are under explored on deforming dynamic articulated actors.
%
In this paper, we propose \emph{TAVA}, a method to create \emph{T}emplate-free \emph{A}nimatable \emph{V}olumetric \emph{A}ctors, based on neural representations.
We rely solely on multi-view data and a tracked skeleton to create a volumetric model of an actor, which can be animated at the test time given novel pose.
Since TAVA does not require a body template, it is applicable to humans as well as other creatures such as animals.
Furthermore, TAVA is designed such that it can recover accurate dense correspondences, making it amenable to content-creation and editing tasks.
Through extensive experiments, we demonstrate that the proposed method generalizes well to novel poses as well as unseen views and showcase basic editing capabilities. The code is available at \url{https://github.com/facebookresearch/tava}
\vspace*{-1em}
\end{abstract}

\section{Introduction}
\label{sec:intro}
Ever since the first 3D vector graphics games in the 1980s, we are striving to build better representations of 3D objects and humans.
With increasing processing power, we can afford to capture, reconstruct and encode increasingly realistic representations.
This makes exploring neural representations for graphical objects particularly appealing---it is a representation that has proven powerful~\cite{mildenhall2020nerf,yu2021plenoctrees,yu2021plenoxels,li2020learning,raj2021anr}, even though still being in its infancy.
%
Recent methods for neural 3D representations go beyond capturing plain texture by modeling radiance fields~\cite{mildenhall2020nerf,barron2021mip,barron2022mipnerf360}, achieving more photo-realistic results than rasterization-based approaches~\cite{zhi2020texmesh,liu2020general,li2020monocular,saito2019pifu}.
However, it is unclear how their representational power can be used to not only capture \emph{static}, but also \emph{dynamic} scenes that can be animated in a meaningful way, making the representations useful for capturing actors that can be ``driven'' post-capture. 
However, due to the high-dimensional nature of pose configurations, 
it is generally neither possible nor practical to capture all pose variations in one capture.
This poses a new problem absent in static settings: \emph{generalization} to out-of-distribution (OOD) poses.
Furthermore, the neural 3D representation is desired to be \emph{editable} as the classical representation like mesh and textures. Some works explored this aspects in the static settings~\cite{yu2021plenoctrees,yu2021plenoxels}, but it is not clear on how to edit neural representations on dynamic actors.

\begin{figure}[!t]
\centering
\includegraphics[width=0.95\linewidth]{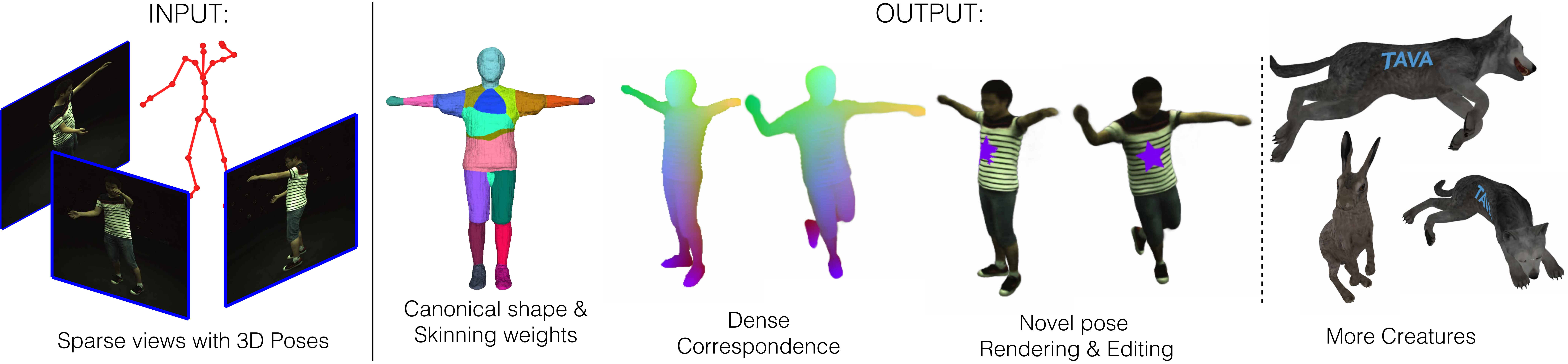}
\vspace*{-1em}
\caption{\textit{Method Overview.} \textbf{Left:} TAVA creates a virtual actor from multiple sparse video views as well as 3D poses. The same skeleton can later be used for animation. \textbf{Center:} TAVA uses this information to create a canonical shape and a pose-dependent skinning function and establishes correspondences across poses. The resulting model can be used for rendering and posing the virtual character as well as editing it. \textbf{Right:} the method can directly used for other creatures as long as a 3D skeleton can be defined.
}
\vspace{-1.5em}
\label{fig:teaser}
\end{figure}

In this paper, we propose TAVA, a novel approach for \textbf{T}emplate-free \textbf{A}nimatable \textbf{V}olumetric \textbf{A}vatars (illustrated in Fig.~\ref{fig:teaser}).
We propose to use coordinate-based radiance fields to capture appearance, leading to high quality, faithful renderings.
We extend the radiance capture with a carefully designed deformation model: while it requires solely 3D skeleton information at training time, it captures non-linear pose-dependent deformations and exhibits stable generalization behavior to unseen poses thanks to being anchored in an LBS formulation.
The radiance field and deformation model are optimized jointly and end-to-end, leading to a simple-to-use and powerful representation: creating it requires only a tracked skeleton and multi-view photometric data, \textbf{no} template mesh or artist-designed rigging; the appearance and the deformation model can complement each other for highest quality results.
These designed properties make TAVA suitable for content creation and editing as well as correspondence-based matching.

In our experiments, we demonstrate that the proposed approach outperforms state-of-the art approaches for animating and rendering human actors on the ZJU motion capture dataset~\cite{peng2021neural}.
Thanks to being template-free, our approach is not limited to capturing humans: we present a detailed evaluation and ablation study on two synthetically rendered animals.
This demonstrates the flexibility of the proposed approach and allows us to show additional applications in content-creation and editing.

\section{Related Work}
\label{sec:relatedwork}

\vspace{1.0em}
\noindent \textbf{Deformable Neural Scene Representations:}
Coordinate-based neural scene representations produce impressive results in encoding shape~\cite{mescheder2019occupancy,chen2019learning,park2019deepsdf} and appearance~\cite{lombardi2019neural,mildenhall2020nerf,sitzmann2019scene}.
These methods train a coordinate-based neural network to model various properties of a scene, e.g., occupancy~\cite{mescheder2019occupancy}, distance to the closest surface~\cite{park2019deepsdf}, or density and color~\cite{mildenhall2020nerf}.
However, making implicit scene representations deformable and animatable remains a challenging research problem.
Nerfies~\cite{park2021nerfies} and Neural 3D Video Synthesis~\cite{li2021neural} handle changes in the scene by optimizing a deformation field and a latent code for each frame.
HyperNeRF~\cite{park2021hypernerf} extends this by additionally creating a hyper-space which allows topology changes of the scene.
Non-Rigid Neural Radiance Fields~\cite{tretschk2021nonrigid} optimize a rigidity model in addition to a deformation field. 
While these methods produce impressive results on dynamic scenes, they are designed to only memorize the scene and cannot control the scene beyond interpolations.

\vspace{1.0em}
\noindent \textbf{Animatable Neural Radiance Fields:}
Recently, many approaches for controllable animatable NeRFs have been proposed.
Neural Actor~\cite{liu2021neural} uses a pose-dependent radiance field by warping rays into the canonical space of a template body model while using 2D texture maps to model fine detail.
NeuralBody~\cite{peng2021neural} anchors latent codes on the vertices of a deformable mesh controlled by LBS. 
The follow-up work Animatable-NeRF~\cite{peng2021animatable} establishes a transformation between view and canonical space through optimizing the inverse deformation field.
Other works like NARF, A-NeRF~\cite{noguchi2021neural,su2021nerf} predict the radiance field at a given 3D location based on its relative coordinates to the bones.
Most recently, a concurrent work HumanNeRF~\cite{weng2022humannerf} produces a free-viewpoint rendering of a human by modeling the inverse deformation as a mixture of affine fields~\cite{lombardi2019neural}.
Yet many of these methods~\cite{noguchi2021neural,su2021nerf} do \emph{not} have a 3D canonical space that preserves correspondences across different poses, which is required for content-creation or editing.
Some~\cite{liu2021neural,peng2021animatable,peng2021neural,weng2022humannerf} are built on top of the SMPL~\cite{loper2015smpl} body template, which prohibits them to be applied to creatures beyond humans. 
Moreover, most of the aforementioned methods either introduce latent codes to better memorize the seen poses~\cite{su2021nerf,peng2021animatable,peng2021neural}, or represent the deformation in the inverse direction from view space to the canonical space~\cite{noguchi2021neural,su2021nerf,peng2021animatable,weng2022humannerf}. Thus they do not generalize well to the unseen poses because the existence of pose-conditioned MLPs. In contrast, our approach is template-free, enables editing, and is designed to be robust to unseen poses.
We provide an overview of the comparison between our method and those previous works in Tab.~\ref{tab:check_check}.
\setlength{\tabcolsep}{4pt}
\begin{table}[!t]
\begin{center}

\begin{adjustbox}{width=0.9\textwidth}
\begin{tabular}{l M{2cm} M{2cm} M{2cm} c}
\hline\hline\noalign{\smallskip}
Methods & Template-free & No Per-frame Latent Code & 3D Canonical Space & Deformation \\
\hline
\noalign{\smallskip}
NARF~\cite{noguchi2021neural} & \cmark & \cmark & \xmark & Inverse \\
A-NeRF~\cite{su2021nerf} & \cmark & \xmark & \xmark & Inverse \\
Animatable-NeRF~\cite{peng2021animatable} & \xmark & \xmark & \cmark & Inverse  \\
HumanNeRF~\cite{weng2022humannerf} & \xmark & \cmark & \cmark & Inverse \\
NeuralBody~\cite{peng2021neural} & \xmark & \xmark & \hspace*{0.15cm}\cmark$^\dagger$ & Forward \\
\hline\noalign{\smallskip}
Ours (TAVA) & \cmark & \cmark & \cmark & Forward \\
\hline\hline
\end{tabular}
\end{adjustbox}
\end{center}
\vspace*{-0.8em}
\caption{\textit{Design differences.} TAVA's use of a forward deformation model without using per-frame latent codes ensures robustness to out-of-distribution poses. Being template-free extends its use to creatures beyond humans. TAVA also allows for content-creation and editing by using a 3D canonical space. $^\dagger$Note that NeuralBody's canonical space consists of the body template \emph{without} clothing.}
\label{tab:check_check}
\vspace*{-3em}
\end{table}
\setlength{\tabcolsep}{1.4pt}

\vspace{1.0em}
\noindent \textbf{Animatable Shapes:}
Non-rigid shape reconstruction often utilizes a canonical space that is fixed across frames, with a deformation model to create a mapping between the canonical and the deformed space.
Traditionally, this has been achieved by extracting a low dimensional articulated mesh ~\cite{borshukov2005universal,carranza2003free,casas20144d,starck2007surface,de2008performance,xu2011video,li2012temporally,li2016spa,volino2014optimal,collet2015high,guo2019relightables,anguelov2005scape}, such as SMPL~\cite{loper2015smpl}, or by extracting a rigged mesh via post-processing.
Several methods~\cite{james2005skinning,hasler2010learning,loper2015smpl,osman2020star,hasler2010learning,jiang2020disentangled,zhou2020unsupervised,xu2020rignet} have been proposed to optimize blend weights and rigs from data.
ARCH~\cite{huang2020arch} deforms an estimated implicit representation to fit to a clothed human using a single image.
Recent approaches model inverse deformation fields~\cite{deng2020nasa,niemeyer2019occupancy,park2021nerfies,pumarola2021d,saito2021scanimate}, which map points from pose-dependent global space to pose-independent canonical space where the surface is represented.
For example, SCANimate~\cite{saito2021scanimate} regularizes the inverse skinning by using a cycle consistency loss.
The main drawback of these inverse deformation approaches is that the inverse transformation is pose dependent and may not generalize well to previously unseen poses.
SNARF~\cite{chen2021snarf} addresses this by learning a forward deformation field instead, mapping points from canonical to pose-dependent deformed space.
However, unlike our appraoch, these methods require 3D geometry supervision and most do not optimize for appearance.

\section{Method}
\label{sec:method}

\begin{figure}[!t]
\centering
\includegraphics[width=0.95\linewidth]{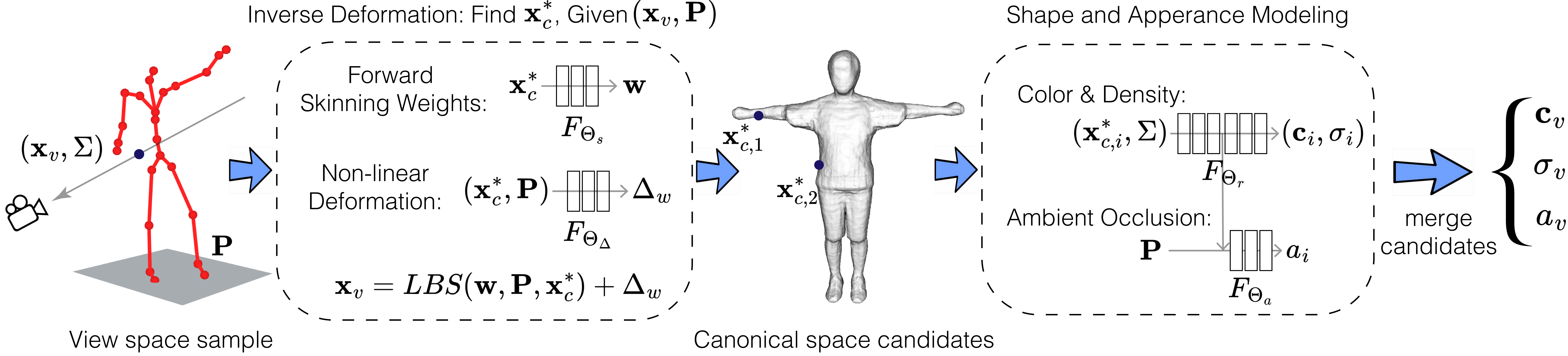}
\vspace*{-0.8em}
\caption{\textit{TAVA Overview}. We use volumetric rendering techniques to create the actor representation. For each sampled point, we use LBS based non-linear deformation combined with a blending weight model for which we identify the root in the canonical space. In this space, we use a color, density, and ambient occlusion model to parameterize the appearance. 
}
\label{fig:overview}
\vspace*{-1.5em}
\end{figure}

Our goal is to create an animatable neural actor from multi-view images with known 3D skeleton information without requiring a body template.
Similar to a traditional personalized body rig, we want to build a representation that not only represents the shape and appearance of the actor but also allows to animate it while maintaining correspondence among different poses and views.
\emph{TAVA} is designed to achieve the above goals with three components: (1)~a canonical representation of the actor in neutral pose, (2)~deformation modeling based on forward skinning, and (3)~volumetric neural rendering with pose-dependent shading. 
Fig.~\ref{fig:overview} illustrates an overview of our method.
To employ volumetric neural rendering in the view space, our method first deforms the samples along a ray back to the canonical space through inverting the forward skinning via root-finding, then queries their colors and densities in the canonical space, as well as the pose-dependent effects. 
Below, we first establish preliminaries, then discuss each of the components.

\subsection{Preliminary: Rendering Neural Radiance Fields}
\label{method:prelim}

NeRF~\cite{mildenhall2020nerf} is a groundbreaking technique for novel view synthesis of a \textit{static} scene.
It models the geometry and view-dependent appearance of the scene by using a multi-layer perceptron (MLP).
Given a 3D coordinate $\mathbf{x} = (x, y, z)$ and the corresponding viewing direction $(\theta, \phi)$, NeRF queries the emitted color $\mathbf{c} = (r, g, b)$ and material density $\sigma$ at that location using the MLP.
A pixel color $C(\mathbf{r})$ can then be computed by accumulating the view-dependent colors along the ray $\mathbf{r}$, weighted by their densities:
\begin{align}
\label{equ:nerf}
    C(\mathbf{r}) = \sum_{i=1}^{N} T_i (1 - \text{exp}(\sigma_i\delta_i))\mathbf{c}_i, \quad \text{where } T_i = \text{exp}(-\sum_{j=1}^{i-1}\sigma_j\delta_j)~,
\end{align}
where $\delta_i$ denotes the distances between the sample points along the ray.
To further take the size of the pixels into consideration, Mip-NeRF~\cite{barron2021mip} extends NeRF to represent each ray $\mathbf{r}$ that passes through a pixel as a cone, and the samples $\mathbf{x}$ along the ray as conical frusta, which can be modeled by multivariate Gaussians $(\bm{\mu}, \bm{\mathrm{\Sigma}})$.
Thus, the density $\sigma$ and view-dependent emitted color $\mathbf{c}$ for a sample on the ray are given by $F_\Theta: (\bm{\mu}, \bm{\mathrm{\Sigma}}, \theta, \phi) \rightarrow (\mathbf{c}, \sigma)$,
where $\bm{\mu} = (x, y, z)$ is the center of the Gaussian and $\bm{\mathrm{\Sigma}} \in \mathbb{R}^{3\times3}$ is its covariance matrix. The loss for optimizing the network parameters $\Theta$ of the neural radiance field is applied between the rendered pixel color $C(\mathbf{r})$ and the ground-truth $\hat{C}(\mathbf{r})$ :
\begin{align}
\label{eq:loss_im}
    \mathcal{L}_{im} = \big|\big|C(\mathbf{r}) - \hat{C}(\mathbf{r})\big|\big|^2_2~.
\end{align}
Please refer to the original papers~\cite{mildenhall2020nerf,barron2021mip} for more details.

\subsection{Canonical Neural Actor Representation}
\label{sec:cano}
We represent an articulated subject as a volumetric neural actor in its canonical space.
The representation includes a \emph{Lambertian} neural radiance field $F_{\Theta_{r}}$ to represent the geometry and appearance of this actor, and a neural blend skinning function $F_{\Theta_{s}}$, which describes how to animate the actor:
\begin{align}
\label{eq:cano}
    F_{\Theta_{r}}: (\mathbf{x}_c, \bm{\mathrm{\Sigma}}) \rightarrow (\mathbf{c}, \sigma),
    \qquad
    F_{\Theta_{s}}: \mathbf{x}_c \rightarrow \mathbf{w},
\end{align}
where $\mathbf{c} = (r, g, b)$ is the material color, $\sigma$ is the material density, and $\mathbf{w}$ is the skinning weights to blend all bone transformations for animation.
Similar to Mip-NeRF~\cite{barron2021mip}, we use a multivariate Gaussian ($\mathbf{x}_c \in \mathbb{R}^3, \bm{\mathrm{\Sigma}} \in \mathbb{R}^{3\times3}$) to estimate the integral of samples within the volume of the discrete samples. Note that in most of the cases, an articulated actor is a Lambertian object, so we exclude view directions from the input of $F_{\Theta_{r}}$.

\noindent\textbf{Discussion.}
This formulation not only models the \emph{canonical} geometry and appearance of an avatar, but also describes its \emph{dynamic} attributes through the skinning weights $\mathbf{w}$.
Unlike previous works, such as SNARF~\cite{chen2021snarf}, which models a pose-dependent geometry in the canonical space, and NARF~\cite{noguchi2021neural} and A-NeRF~\cite{su2021nerf}, which entirely skips canonical space modeling, our method is based on a canonical representation that fully eliminates \emph{any} effects of pose on the geometry and appearance.
Moreover, the skinning weights learnt in the canonical space remain valid for a large range of poses, meaning that the actor is ready to be animated in novel poses outside of the training distribution (see Sec.~\ref{sec:eval} for validation of its robustness to out-of-distribution novel poses).
Last but not least, our representation eases the correspondence finding problem across different poses and views, because the matching can be done in the pose-independent canonical space (see Sec.~\ref{sec:eval} for results).

\subsection{Skinning-based Deformation}

\subsubsection{Forward Skinning.}
With the skinning weights $\mathbf{w} = (w_1,  w_2, ..., w_B, w_{bg}) \in \mathbb{R}^{B+1}$ defined in the canonical space, and given a pose $\mathbf{P} = \{\mathbf{T}_1, \mathbf{T}_2, ..., \mathbf{T}_B\} \in \mathbb{R}^{B\times4\times4}$, we use forward LBS to define the deformation of a point $\mathbf{x}_c$ in the canonical space to $\mathbf{x}_v $ in the view space:
\begin{align}
    \mathbf{x}_v 
    = LBS(\mathbf{w}(\mathbf{x}_c; \Theta_s), \mathbf{P}, \mathbf{x}_c)
    = \left[ \sum_{j=1}^{B} w_{j}(\mathbf{x}_c ; \Theta_s) \cdot \mathbf{T}_j + w_{bg} \cdot \mathbf{I_d} \right] \mathbf{x}_c,
\label{Eq:lbs}
\end{align}
where $\mathbf{I}_d \in \mathbb{R}^{4\times4}$ is an identity matrix. Similar to~\cite{weng2022humannerf}, we extend the classic LBS 
defined only on the surface geometry of an object to the entire 3D space by introducing an additional term $ w_{bg} \cdot \mathbf{I_d}$. 
This term allows the points in the background and empty space to \emph{not} follow the skeleton when it is deformed.
However, LBS is not sufficient for capturing some of the non-linear deformations, such as muscles and clothing dynamics~\cite{loper2015smpl}.
Thus, we introduce an additional term $F_{\Theta_{\Delta}}: (\mathbf{x}_c, \mathbf{P}) \rightarrow \Delta_w \in \mathbb{R}^3$ on top of the learned LBS to model these deformations:
\begin{align}
    \mathbf{x}_v = LBS(\mathbf{w}(\mathbf{x}_c; \Theta_s), \mathbf{P}, \mathbf{x}_c) + \Delta_w(\mathbf{x}_c, \mathbf{P}; \Theta_\Delta).
\label{Eq:nolinear_lbs}
\end{align}

\subsubsection{Inverse Skinning.} 
To render this model, 
we need to query color and density in the view space. So it is required to find the the correspondence $\mathbf{x}_c$ in the canonical space for each $\mathbf{x}_v$ in the view space.
As our forward skinning in Eq.~\ref{Eq:nolinear_lbs} is defined through neural networks, there is no analytical form for the inverse skinning. So, inspired by SNARF~\cite{chen2021snarf}, we pose this as a root finding problem:
\begin{align}
    \text{Find } \mathbf{x}_c^*, \quad
    \text{s.t. } f(\mathbf{x}_c^*) = LBS(\mathbf{w}(\mathbf{x}_c^*; \Theta_s), \mathbf{P}, \mathbf{x}_c^*) + \Delta_w(\mathbf{x}_c^*, \mathbf{P}; \Theta_\Delta) - \mathbf{x}_v = \mathbf{0}
\end{align}
and solve it numerically using Newton's method:
\begin{align}
    \mathbf{x}_c^{(k+1)} = \mathbf{x}_c^{(k)} - (\mathbf{J}^{(k)})^{-1} f(\mathbf{x}_c^{(k)}),
\label{Eq:newton}
\end{align}
where $\mathbf{J}^{(k)} \in \mathbb{R}^{3\times3}$ is the Jacobian of $f(\mathbf{x}_c^{(k)})$ at the $k$-th step.
Since the inverse skinning might be a one-to-many mapping when there is contact happening between body parts, we initialize Newton's method with multiple candidates using the inverse rigid transformation $\{\mathbf{x}_{c,i}^{(0)}\} = \{\mathbf{T}_i^{-1} \cdot \mathbf{x}_v\}$.
However, simply applying all $B+1$ transformations to initialize the Newton's method would lead to $B+1$ canonical candidates to be processed, making it impractical for volumetric rendering as the complexity grows linearly in $B$. 
As points are less likely to be affected by bones further away, 
we only use the transformations of its $K=5$ nearest bones by measuring the Euclidean distance between the point and the bones in the view space.
This dramatically reduces the computational burden of the root finding process as well as the following canonical querying, making it feasible for neural rendering.
With that, our inverse skinning leads to multiple correspondences for a point in the view space through root finding (r.f.):
\begin{align}
    \mathbf{x}_v \xrightarrow[]{\text{r.f.}} \{\mathbf{x}_{c,1}^*, \mathbf{x}_{c,2}^*, ..., \mathbf{x}_{c,K}^*\}
\end{align}
The gradients of the network parameters $\Theta_s$ and $\Theta_\Delta$ can then be analytically computed for the inverse skinning~\cite{chen2021snarf}:
\begin{align}
    \frac{\partial \mathbf{x}^*_{c,i}}{\partial \Theta_s} = - \left[ \frac{\partial \mathbf{x}_v}{\partial \mathbf{x}^*_{c,i}} \right]^{-1} \left[ \frac{\partial \mathbf{x}_v}{\partial \Theta_s} \right].
    \qquad
    \frac{\partial \mathbf{x}^*_{c,i}}{\partial \Theta_\Delta} = - \left[ \frac{\partial \mathbf{x}_v}{\partial \mathbf{x}^*_{c,i}} \right]^{-1} \left[ \frac{\partial \mathbf{x}_v}{\partial \Theta_\Delta} \right].
\end{align}
Please refer to the supplemental material for their derivations.

\subsection{Deformation-based Neural Rendering}
\label{sec:render}
Similar to Mip-NeRF~\cite{barron2021mip}, we render the color of a pixel by accumulating the samples $(\mathbf{x}_v, \bm{\mathrm{\Sigma}})$ along each pixel ray, using Eq.~\ref{equ:nerf}.
Instead of directly querying the color and density of $\mathbf{x}_v$ in the view space, we first find the point's canonical correspondence candidates using the inverse skinning $\mathbf{x}_v \xrightarrow[]{\text{r.f.}} \{\mathbf{x}_{c,1}^*, \mathbf{x}_{c,2}^*, ..., \mathbf{x}_{c,K}^*\}$, and then query the colors and densities for all those candidates in the canonical space .
\begin{align}
    F_{\Theta_r}: (\mathbf{x}_{c,i}^*, \bm{\mathrm{\Sigma}}) \rightarrow (\mathbf{c}_{i}^*, \sigma_{i}^*).
\end{align}

However, for a dynamic object, the shading on the surface may change depending on pose due to self-occlusion.
This can lead to colors in the view space being darker than the colors in the canonical space, providing inconsistent supervision signals.
However, it is non-trivial to accurately model this self-occlusion without ray tracing  (including secondary rays) and known global illumination. 
A simple but effective estimator, widely used in modern rendering engines like Unreal and Blender is ambient occlusion, in which the shading caused by occlusion is modeled by \emph{a scaling factor} multiplied with the color values, where the value is calculated by the percentage of view directions being occluded around each point on the surface.
Since it is an attribute defined at each coordinate that depends on the global geometry of the actor, we model this shading effect use a coordinate based MLP $F_{\Theta_a}$ conditioned on the pose $\mathbf{P}$ of the actor:
\begin{align}
    F_{\Theta_r}: (\mathbf{x}_{c,i}^*, \bm{\mathrm{\Sigma}}) \rightarrow \mathbf{h} \rightarrow (\mathbf{c}_{i}^*, \sigma_{i}^*),
    \qquad
    F_{\Theta_a}: (\mathbf{h}, \mathbf{P}) \rightarrow a_i^*,
\end{align}
where $\mathbf{h}$ is an intermediate activation from $F_{\Theta_r}$, and $a_i^*$ is the ambient occlusion at this location under pose $\mathbf{P}$.
Note that only the ambient occlusion $a_i^*$ is pose-conditioned, which makes sure the actor (geometry and appearance) is represented in a canonical space that is pose-independent, as described in Sec.~\ref{sec:cano}.

With $(\mathbf{c}^*_i, \sigma^*_i, a^*_i)_{i=1, ..., K}$ queried in the canonical space, we then need to merge the $K$ candidates to get the final attributes $(\mathbf{c}_v, \alpha_v, a_v)$ for the sample $(\mathbf{x}_v, \bm{\mathrm{\Sigma}})$ in the view space.
In the case of articulated objects, where multiple canonical point may originate from the same location, the one with the maximum density would dominate that location.
Similar to previous works~\cite{chen2021snarf,deng2020nasa}, 
we choose the attributes of $\mathbf{x}_v$ from all canonical candidates based on their density:
\begin{align}
    \mathbf{c}_v = \mathbf{c}_{c, t}^* \quad \sigma_v = \sigma_{c, t}^* \quad a_v = a_{c, t}^*
    \qquad \text{where } t = \argmax_{i}(\{\sigma_{c,i}^*\}),
\end{align}
then we use $(\mathbf{c} = a_v * \mathbf{c}_v, \sigma = \sigma_v)$ as the final emitted color and density in the view space, for the volumetric rendering in Eq.~\ref{equ:nerf}.

Note that in general there is no way to guarantee that the inverse root finding converges. 
In practice, root finding fails for 1\% to 8\% 
of the points in the view space, making it impossible to query their attributes. 
For these points, an option is to just simply set their densities to zero, which would only be problematic if the points are close to the surface.
A slightly better way is to estimate the color and density for those points by interpolating the attributes from their nearest valid neighbors along the ray.
We conduct experiments on both strategies in Sec.~\ref{sec:eval}, which results in slightly better performance.
We choose the second strategy in our full model.

\subsection{Establishing Correspondences}
\label{sec:corr}
As our method is endowed with a 3D canonical space, we have the ability to trace surface correspondences across different views and poses.
When rendering an image using Eq.~\ref{equ:nerf}, besides accumulating colors $\{\mathbf{c}_m\}$ of the samples $\{\mathbf{x}_{v,m}\}$ along the ray $\mathbf{r}$, we also accumulate the corresponding canonical coordinates $\{\mathbf{x}_{c,m}\}$:
\begin{align}
    X(\mathbf{r}) = \sum_{m=1}^{N} T_m (1 - \text{exp}(\sigma_m\delta_m))\mathbf{x}_{c,m},
\end{align}
where $X(\mathbf{r}) \in \mathbb{R}^3$ is the coordinate in the canonical space that corresponds to the pixel of the ray.
With that, for images under different poses / views, we can compute their \emph{dense} pixel-to-pixel correspondences by matching $X(\mathbf{r})$ in the canonical space using the nearest neighbor algorithm.

\subsection{Training loss}

Besides the image loss $\mathcal{L}_{im}$ defined in Eq.~\ref{eq:loss_im}, we also employ two auxiliary losses that help the training.
Due to the fact that all the points along a bone should have the same transformation, we encourage the skinning weights $\mathbf{w}$ of samples $\mathbf{\Bar{x}}_c$ on the bones to be one-hot vectors $\mathbf{\hat{w}}$ (noted as $\mathcal{L}_{w}$).
We also encourage the non-linear deformations $\mathbf{\Delta}_v$ of those samples to be zero given any pose $\mathbf{P}$ (noted as $\mathcal{L}_{\Delta}$).
We use $MSE$ to calculate both, $\mathcal{L}_{w} = ||\mathbf{w}(\mathbf{\Bar{x}}_c) - \mathbf{\hat{w}})||^2_2 $ and $\mathcal{L}_{\Delta} = ||\Delta_w(\mathbf{\Bar{x}}_c, \mathbf{P}) - \mathbf{0})||^2_2$.
Our final loss is:
$\mathcal{L} = \mathcal{L}_{im} + \lambda\mathcal{L}_{w} + \beta\mathcal{L}_{\Delta},$
where $\lambda$ is set to $1.0$ and $\beta$ is set to $0.1$ in all our experiments. 
\section{Experiments}
\label{sec:experiments}
\begin{figure}[!t]
\centering
\includegraphics[width=0.9\linewidth]{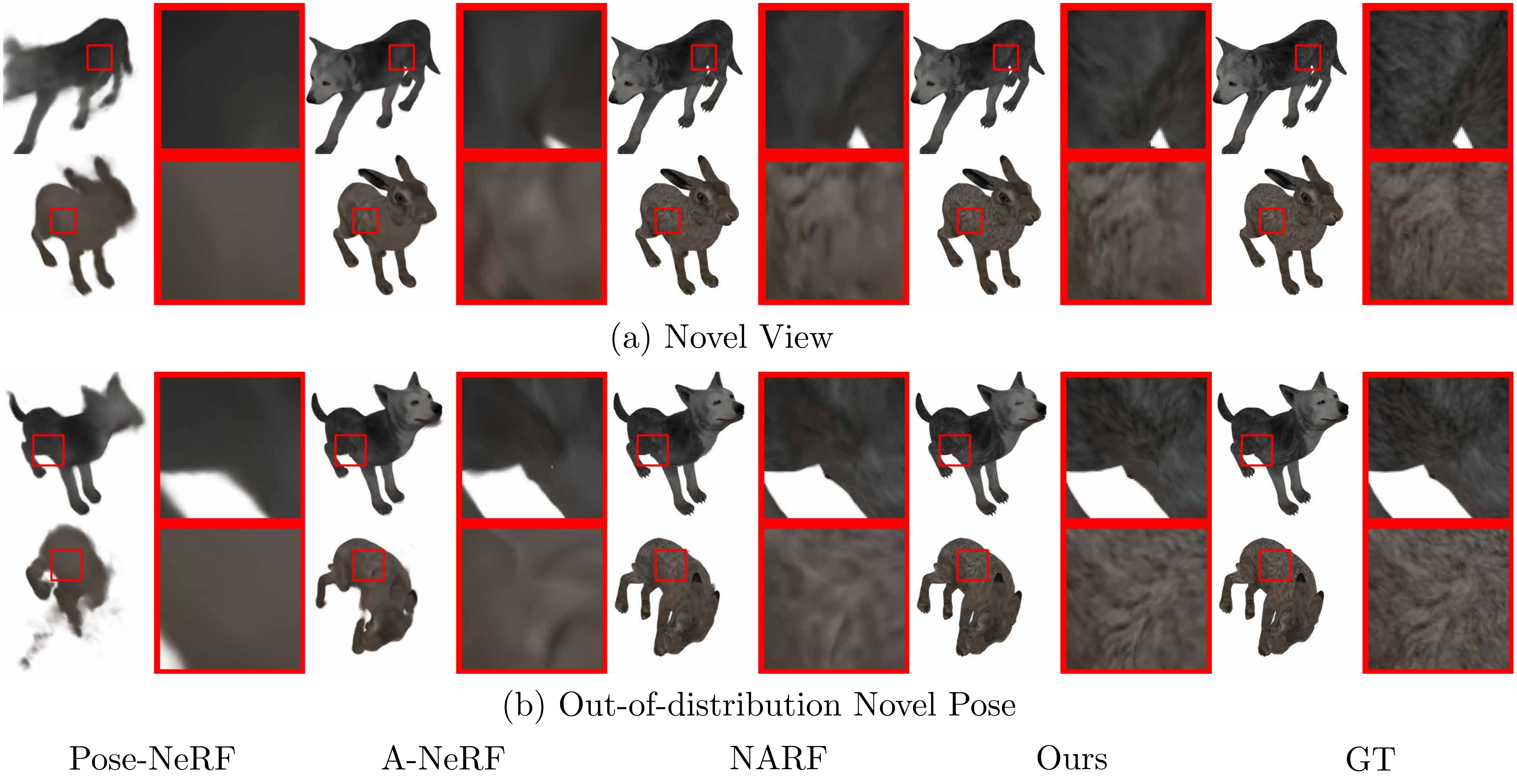}
\vspace*{-1.5em}
\caption{\textit{Comparison with template-free methods on the Hare and Wolf subjects.}
}
\vspace*{-1.5em}
\label{fig:comparison_animal}
\end{figure}

\subsection{Datasets}
We conduct experiments on 1)~four human subjects (313, 315, 377, 386) in the ZJU-Mocap dataset~\cite{peng2021neural}, a public multi-view video dataset for human motion, and 2)~two synthetic animal subjects (Hare, Wolf) introduced in this paper, rendered from multiple views using Blender.

\vspace{.5em}
\noindent\textbf{Data Splits.}
Prior works~\cite{peng2021animatable,peng2021neural} create the \emph{train} and \emph{val} sets on the ZJU-Mocap dataset by simply splitting each video with $500\sim2200$ frames into two splits, where the \emph{training} set has $60\sim300$ frames and the \emph{validation} set has $300\sim1000$ frames.
This is not an ideal split to evaluate pose synthesis performance because 1)~a training set with $60$ consecutive frames in a $30$fps video does not sufficiently cover pose variation to learn from, and 2)~due to the repetitive motion of the actors, 
quite often similar poses are in both the \emph{training} and \emph{validation} sets, which should be avoided for evaluating a method on pose generalization.
Therefore we establish a new protocol to split the dataset by clustering the frames based on pose similarity.
Specifically, for each subject, we first randomly withhold a chunk of consecutive frames to be the \emph{test} set, for the purpose of the final evaluation.
Then, we use the K-Medoids algorithm on the remaining frames to cluster them into $K=10$ clusters, based on pose similarity measured by the V2V Euclidean distance using ground-truth mesh. 
The most different cluster is selected as the \emph{$\text{val}_{\text{pose}}^{\text{ood}}$} set, in which the frames are all considered to contain the \emph{out-of-distribution} poses.
For the remaining $9$ clusters, we randomly split each cluster $2:1$ to form \emph{train} and \emph{$\text{val}_{\text{pose}}^{\text{ind}}$} sets, where the frames in \emph{$\text{val}_{\text{pose}}^{\text{ind}}$} still contains new poses which are considered to be \emph{in the distribution} of the training set. For the view splits, we follow the protocol from~\cite{peng2021animatable,peng2021neural} for ZJU-Mocap, where $4$ views are used for training and $17$ views for testing. The animal subjects have $10$ random views for training and $10$ for testing. We denote \emph{$\text{val}_{\text{view}}$} as our novel-view synthesis evaluation set, which contains all the training poses but rendered from different viewpoints.
Please see the supplemental material for more details.

\setlength{\tabcolsep}{4pt}
\begin{table}
\vspace*{-1.5em}
\begin{center}
\resizebox{0.9\linewidth}{!}{
\begin{tabular}{l cc cc cc}
\toprule
& \multicolumn{2}{c}{Novel-view} & \multicolumn{2}{c}{Novel-pose (ind)} & \multicolumn{2}{c}{Novel-pose (ood)} \\
& PSNR $\uparrow$ & SSIM $\uparrow$ & PSNR $\uparrow$ & SSIM $\uparrow$ & PSNR $\uparrow$ & SSIM $\uparrow$ \\
\midrule
\noalign{\smallskip}
\emph{\underline{\textbf{SMPL-based Methods}}} & & & & & & \\
Animatable-NeRF~\cite{peng2021animatable} & 30.75 & 0.971 & 29.34 & 0.966 & 28.67 & 0.961 \\
NeuralBody~\cite{peng2021neural} & \textbf{33.91} & \textbf{0.983} & \textbf{33.43} & \textbf{0.984} & \textbf{30.33} & \textbf{0.969} \\
\emph{\underline{\textbf{Template-free Methods}}} & & & & & & \\
Pose-NeRF & 31.88 & 0.975 & 32.09 & 0.976 & 28.43 & 0.954 \\
A-NeRF~\cite{su2021nerf} & 32.45 & 0.978 & 32.65 & 0.978 & 30.41 & 0.967 \\
NARF~\cite{noguchi2021neural} & 32.94 & 0.980 & 33.21 & 0.980 & 30.60 & 0.968 \\
\hline\noalign{\smallskip}
Ours & \textbf{33.11} & \textbf{0.981} & \textbf{33.35} & \textbf{0.981} & \textbf{30.69} & \textbf{0.969} \\
\bottomrule
\end{tabular}
}
\end{center}
\vspace*{-0.8em}
\caption{\textit{Comparisons on the ZJU Mocap subjects}. We compare with both, template-free and template-based  methods.}
\vspace*{-5em}
\label{tab:comparison_zju}
\end{table}
\setlength{\tabcolsep}{1.4pt}
\setlength{\tabcolsep}{4pt}
\begin{table}
\begin{center}
\begin{adjustbox}{width=0.9\textwidth}
\begin{tabular}{l cc ccc ccc}
\toprule
& \multicolumn{2}{c}{Novel-view} & \multicolumn{3}{c}{Novel-pose (ind)} & \multicolumn{3}{c}{Novel-pose (ood)} \\
& PSNR $\uparrow$ & SSIM $\uparrow$  & PSNR $\uparrow$ & SSIM $\uparrow$ & P2P $\downarrow$ & PSNR $\uparrow$ & SSIM $\uparrow$ & P2P $\downarrow$ \\
\midrule
\noalign{\smallskip}
Pose-NeRF & 23.40 & 0.974 & 21.93 & 0.941 & 197.11 & 16.62 & 0.925 & 88.85  \\
A-NeRF~\cite{su2021nerf} & 31.26 & 0.976 & 31.22 & 0.977 & 31.52 & 25.66 & 0.967 & 19.04\\
NARF~\cite{noguchi2021neural} & 36.55 & 0.988 & 36.65 & 0.988 & 9.28 & 30.92 & 0.982 & 8.46 \\
\hline\noalign{\smallskip}\textbf{}
Ours & \textbf{37.30} & \textbf{0.991} & \textbf{37.45} & \textbf{0.991} & \textbf{4.30} & \textbf{35.77} & \textbf{0.990} & \textbf{3.38} \\
\bottomrule
\end{tabular}
\end{adjustbox}
\end{center}
\vspace*{-0.8em}
\caption{\textit{Comparisons on the animal subjects}. P2P is pixel-to-pixel error, for measuring image correspondences across different poses.}
\vspace*{-4em}
\label{tab:comparison_animal}
\end{table}
\setlength{\tabcolsep}{1.4pt}

\subsection{Evaluation and Comparison}
\label{sec:eval}
\vspace*{-1em}

\vspace{.5em}
\noindent\textbf{Baselines.}
We compare our work with two types of previous methods: 1)~Template-free methods, including NARF~\cite{noguchi2021neural} and A-NeRF~\cite{su2021nerf}, as well as 2)~SMPL-based methods, including Animatable-NeRF~\cite{peng2021animatable} and NeuralBody~\cite{peng2021neural}.
As our baseline, we use Pose-NeRF: we slightly modify Mip-NeRF~\cite{barron2021mip} to learn the density and color conditioned on pose.
We conduct experiments for all the moethods above on ZJU-Mocap, but exclude Animatable-NeRF and NeuralBody for the animal subjects (they require a  template 3D model).
Although code is available for each method, 
we noticed that each is using a different set of hyper-parameters for neural rendering (e.g., number of MLP layers, number of samples, near and far planes) and different training schedules, all of which are not related to method design but can greatly affect the performance.
To make as-fair-as-possible comparisons, we integrated the template-free methods, NARF and A-NeRF, into our code base, which shares the same set of hyper-parameters\footnote{For NARF, our re-implementation achieves better performance than it's official implementation. Please refer to the supplmental material for further details.}. 
For Animatable-NeRF and NeuralBody, we use the original implementations since their designs are based on the SMPL body template.

\begin{figure}
\centering
\vspace*{-1em}
\includegraphics[width=0.9\linewidth]{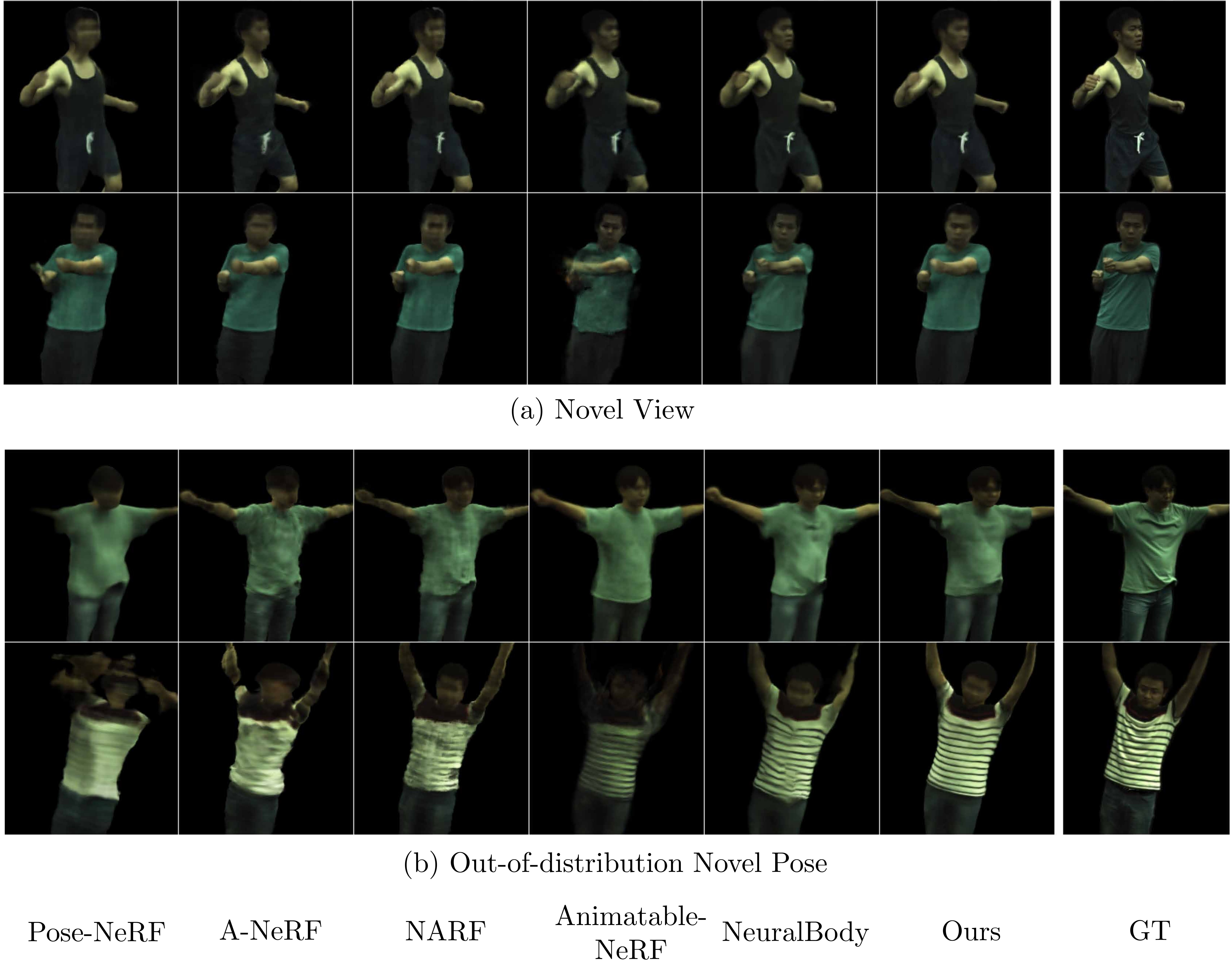}
\vspace*{-1.5em}
\caption{\textit{Rendering quality comparison with all baseline methods on the ZJU-Mocap Dataset.} Note that Animatable-NeRF and NeuralBody rely on the SMPL body model, and the other approaches do not.
}
\vspace*{-2em}
\label{fig:comparison_zju}
\end{figure}

\vspace{0.5em}
\noindent\textbf{Novel-view Synthesis.}
In this task, we conduct experiments on both, the ZJU-Mocap dataset and the two animal subjects Hare and Wolf, using \emph{$\text{val}_{\text{view}}$} set.
As shown in Tab.~\ref{tab:comparison_animal} and Tab.~\ref{tab:comparison_zju}, our method outperforms other 
template-free methods measured by PSNR and SSIM.
On the ZJU-Mocap dataset, our method achieves comparable performance with two template-based methods, Animatable-NeRF and NeuralBody, which greatly benefit from the SMPL body template, but do not work on other creatures like animals.
See Fig.~\ref{fig:comparison_zju} for a qualitative comparison.

\begin{figure}
\centering
\includegraphics[width=0.75\linewidth]{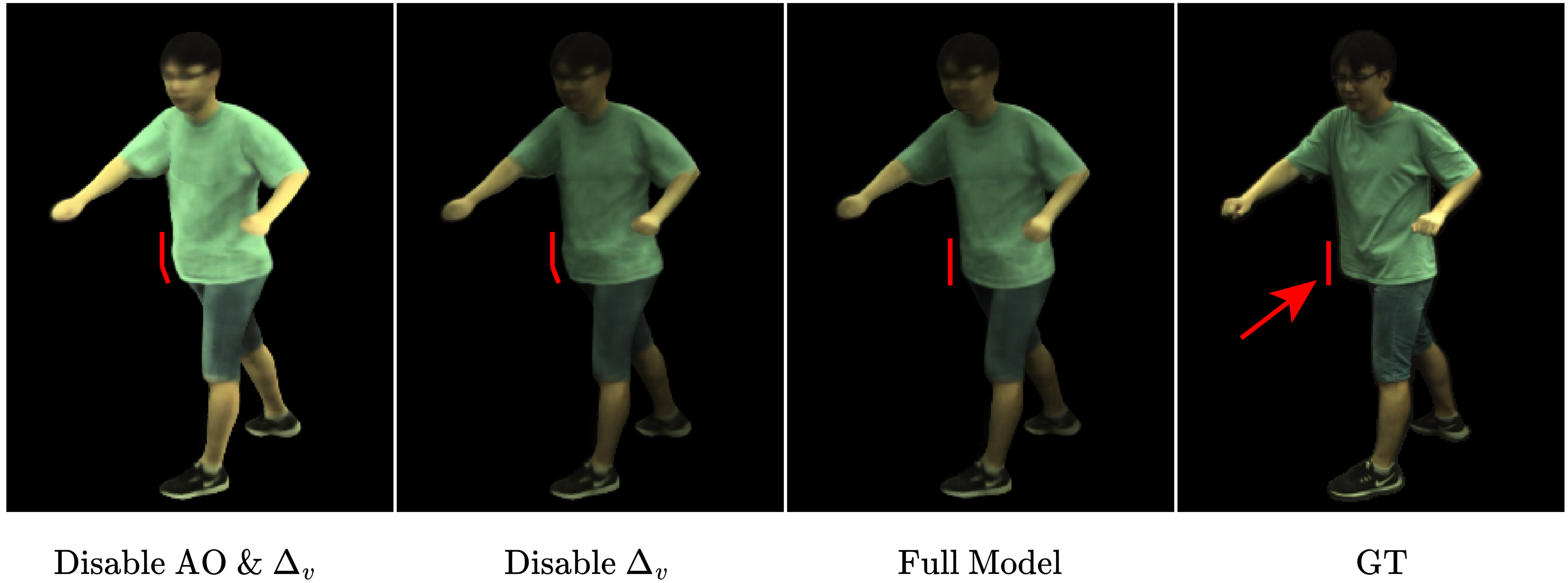}
\vspace*{-1em}
\caption{\textit{Ablation on the ambient occlusion (AO) and non-linear deformation ($\Delta_v$) terms}. Due to our designs, we can train our full model with both enabled, then disable them during inference to ablate their effects. 
}
\vspace*{-1em}
\label{fig:ablation}
\end{figure}

\vspace{0.5em}
\noindent\textbf{Novel-pose Synthesis.}
Due to the high interdependency of appearance changes caused by pose and motion, novel-pose synthesis is a more challenging task than novel-view synthesis, especially for poses that are out of the training distribution.
To carefully study this problem, we conduct experiments on both, in-distribution (InD) novel poses, using the \emph{$\text{val}_{\text{pose}}^{\text{ind}}$} set, and out-of-distribution(OOD) novel poses, using the \emph{$\text{val}_{\text{pose}}^{\text{ood}}$} set.
Our experiments reveal that for InD poses, the performance of nearly all the methods are consistent with their performance on the novel-view task, as shown in Tabs.~\ref{tab:comparison_zju},~\ref{tab:comparison_animal}.
However, there is a huge drop in performance from InD poses to OOD pose ($0.67$db$\sim2.66$db on ZJU-Mocap; $1.68$db$\sim5.73$db on animals).
This is not surprising if the method contains neural networks that directly infer appearance information from pose input: generalization to vastly different pose inputs can not be expected.
One of the main goals in this paper is to reduce this reliance of the neural networks to the pose input, for improving the robustness of the method to the OOD poses.
Our method benefits from explicitly incorporating the \emph{forward} LBS. We observe only an $1.68$db performance drop comparing InD to OOD poses on the animal subjects, whereas other methods suffer from $\sim 5$db performance drops, as shown in Tab.~\ref{tab:comparison_animal}.
Since these two synthetic subjects are not rendered with pose-dependent shading, and do not have ``clothing'' deformations, we disabled the ambient occlusion $a$ (set to $1.$) and non-linear deformation $\Delta_v$ (set to $0.$) terms in our method during both, training and inference.
These synthetic subjects allows us to study the underlying formulation of the articulation deformation, and we show here the \emph{forward} LBS-based deformation is more reliable than the \emph{inverse} deformation used in the baselines which takes pose as input to the MLP.
The results on the ZJU-Mocap dataset in Tab.~\ref{tab:comparison_zju} show that our method outperforms both, the template-free and template-based methods, on the OOD poses.
All the methods are prone to nearly the same drop in performance on the ZJU-Mocap dataset comparing InD to OOD poses.
This is, because currently all of these methods, including ours, are still implicitly modeling pose-dependent shading effects (e.g., self-occlusion) as either a neural network or a latent code during training, which does not generalize well to OOD poses.
Our method, though, provides a possibility to factor out the shading effects during inference and reveal the albedo color of the actor, which yields better generalization but is not suitable for evaluation comparing to the ground-truth, as shown in Fig.~\ref{fig:ablation}.
See Figs.~\ref{fig:comparison_animal},~\ref{fig:comparison_zju} for qualitative comparisons.

\begin{figure}
\centering
\vspace*{-1.5em}
\includegraphics[width=0.9\linewidth]{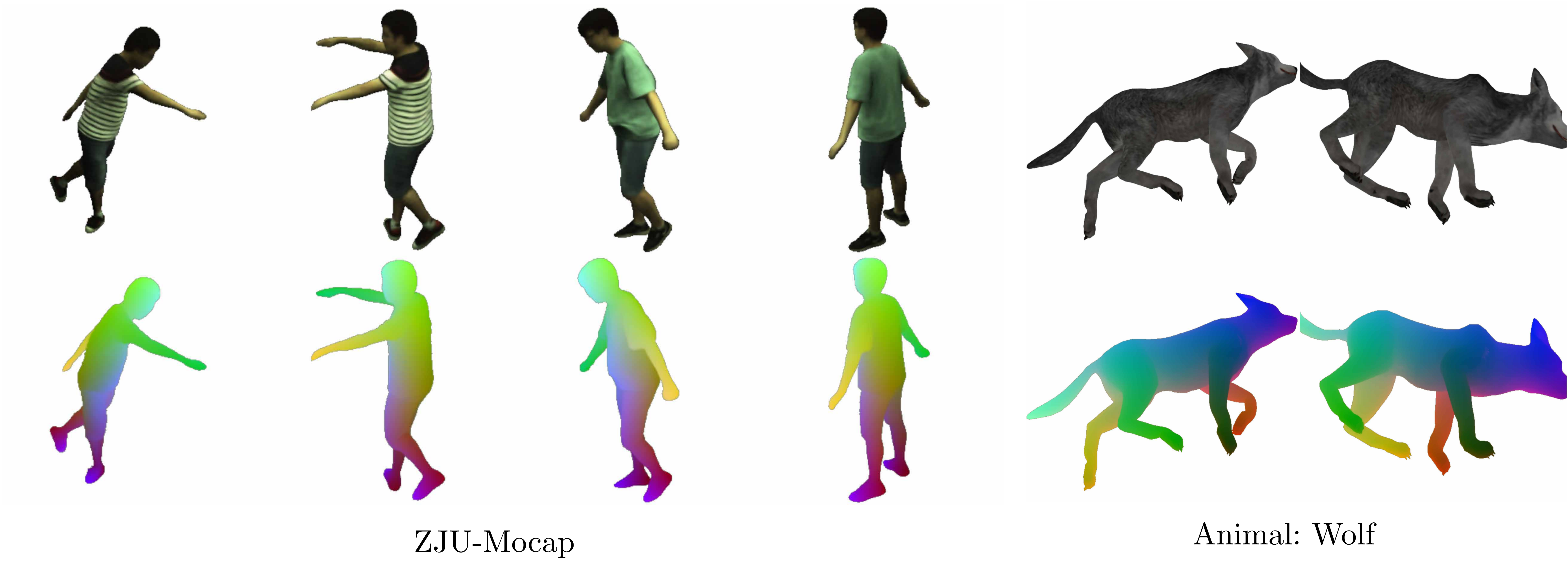}
\vspace*{-1em}
\caption{\textit{Rendering with Dense Correspondence}. We show results of our novel-view rendering with dense correspondences. On the ZJU Mocap dataset, correspondences across different subjects can also be built because they share the same canonical pose. 
}
\vspace*{-2.0em}
\label{fig:demo_corr}
\end{figure}

\begin{figure}
\centering
\includegraphics[width=0.9\linewidth]{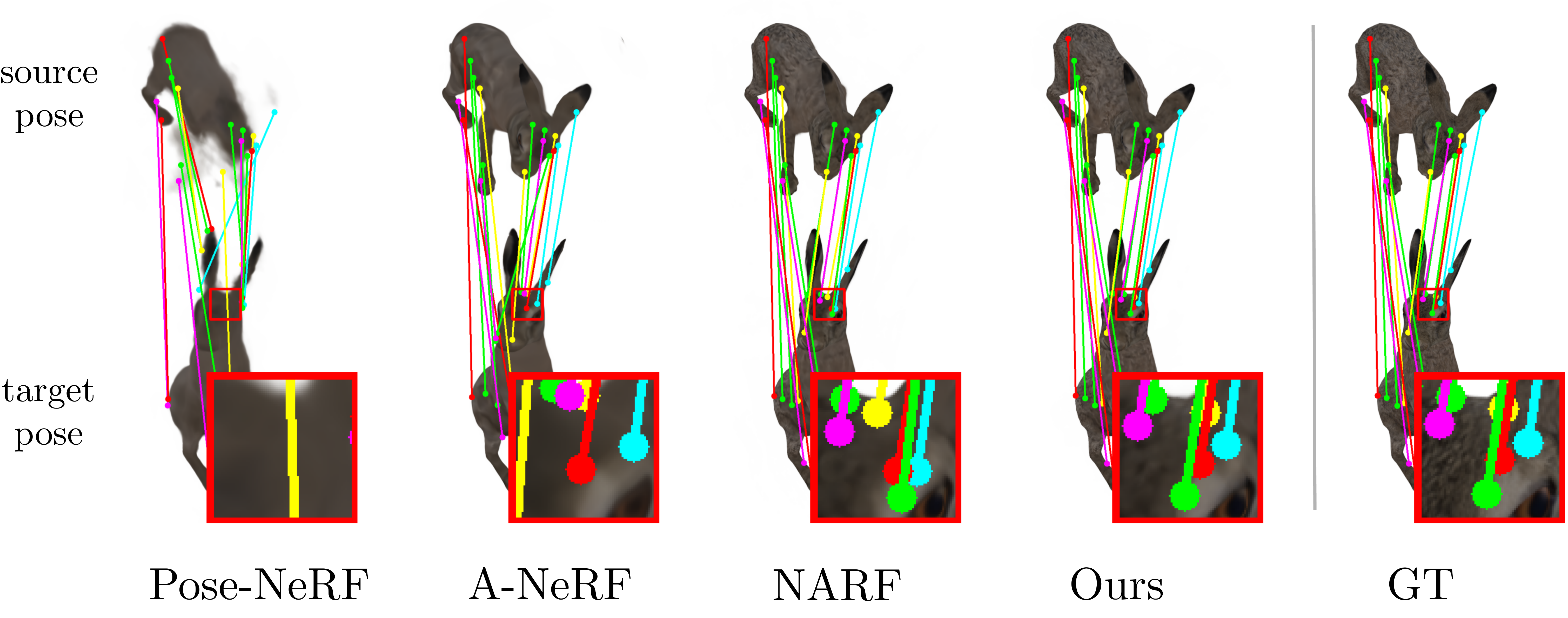}
\vspace*{-1.5em}
\caption{\textit{Correspondence comparison on Hare}. We find the correspondence for the same set of pixels in the source image in the target image.
Both source and target are rendered in novel poses.
}
\vspace*{-2em}
\label{fig:comparison_corr}
\end{figure}

\begin{figure}
\centering
\vspace*{-2em}
\includegraphics[width=0.95\linewidth]{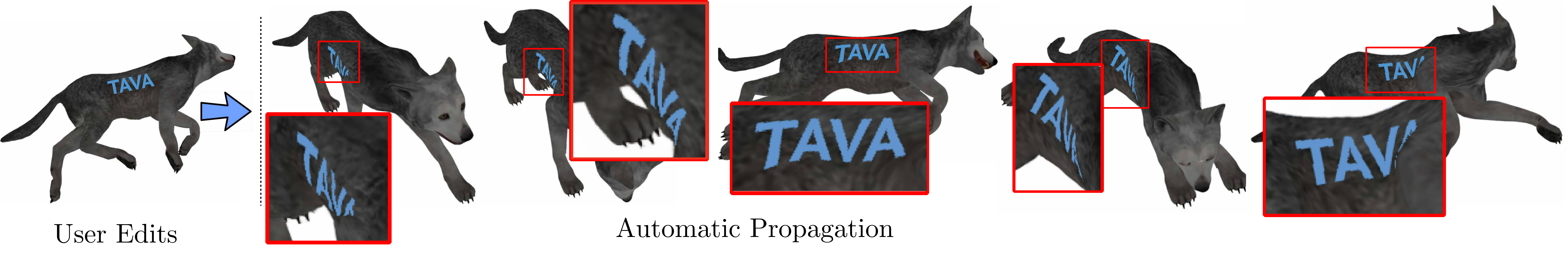}
\vspace*{-1em}
\caption{\textit{Rendering $\&$ Editing}. We show results of our novel-pose rendering with content editing. We manually attach a logo to the image on the left, then use our pixel-to-pixel correspondences to automatically propagate the logo to different poses $\&$ views. 
}
\vspace*{-3.5em}
\label{fig:demo_editing}
\end{figure}

\vspace{0.5em}
\noindent\textbf{Pixel-to-Pixel Correspondences.}
We quantitatively evaluate correspondence on the animal subjects against 
Pose-NeRF, A-NeRF~\cite{su2021nerf} and NARF~\cite{noguchi2021neural}.
We show qualitative results on ZJU-Mocap since no ground-truth correspondences are available (see Fig.~\ref{fig:demo_corr}).
Even though neither of the baseline methods demonstrate that they can establish correspondences, we still tried our best to create a valid comparison\footnote{Pose-NeRF, A-NeRF and NARF all query the color and density of ${(\mathbf{x}_v, \mathbf{P})}$ in a higher dimensional ($>3$) space, where we do the nearest neighbor matching for them using our approach as described in Sec.~\ref{sec:corr}. Please refer to the supp.mat. for further details.}.
For quantitative evaluation, we randomly sample $2000$ image pairs ($A, B$) in \emph{$\text{val}_{\text{pose}}^{\text{ood}}$} set, and use the ground truth mesh to establish ground-truth pixel-to-pixel correspondences $(\chi_A \rightarrow \chi_B)$ for every pair of images ($A \rightarrow B$), where $\chi_A$ and $\chi_B$ are the corresponding image coordinates.
Then, we use each method to render this pair of images, and find the correspondences of $\chi_A$ in $B$ as $\chi^*_B$.
The pixel-to-pixel error (P2P) is then calculated as the average distance between $\chi_B$ and $\chi^*_B$: $\text{P2P} = ||\chi_B - \chi^*_B||^2_2$.
As shown in Tab.~\ref{tab:comparison_animal} and Fig.~\ref{fig:comparison_corr}, our method achieves over 2x more accurate correspondences ($3.38$px v.s. $8.46$px error in a $800\times800$ image) compared to the baselines.
We visualize the extracted dense correspondences of our method in Fig.~\ref{fig:demo_corr}, which shows that correspondences across different subjects can be established as long as they share the same canonical pose (T-pose in ZJU-Mocap).
Further more, we demonstrate that accurate correspondences can be used for content editing in Fig.~\ref{fig:demo_editing}.

\subsection{Ablation Studies.}

Thanks to our model design, we can train a full model with the non-linear deformation $\Delta_v$ and the ambient occlusion AO enabled, then strip them out at inference time.
Fig.~\ref{fig:ablation} shows a qualitative result to visually demonstrate the impacts on the full model.
Notice that without AO, the shading effects are removed during rendering, which produces overall brighter images than the ground-truth.
This is an expected effect, but prohibits quantitative evaluation.
Furthermore, we ablate these two design decisions during training.
For the non-linear deformation, our ablation is to simply disable it during training to see if the LBS is enough to model the deformation.
For the AO, our ablation is to compare it with predicting a pose-dependent color by conditioning pose to the color branch of $F_{\Theta_r}$, and disabling the AO branch $F_{\Theta_a}$.
As shown in Tab.~\ref{tab:abl}, both design decisions contribute to the final model performance.
Lastly, we show the two different strategies to deal with root finding failures described in Section~\ref{sec:render} in Tab.~\ref{tab:abl}.
Using the interpolation strategy results in slightly better performance.

\setlength{\tabcolsep}{4pt}
\begin{table}
\vspace*{-1em}
\begin{center}
\resizebox{0.9\linewidth}{!}{
\begin{tabular}{l cc cc cc}
\hline\hline\noalign{\smallskip}
& \multicolumn{2}{c}{Novel-view} & \multicolumn{2}{c}{Novel-pose (ind)} & \multicolumn{2}{c}{Novel-pose (ood)} \\
& PSNR $\uparrow$ & SSIM $\uparrow$ & PSNR $\uparrow$ & SSIM $\uparrow$ & PSNR $\uparrow$ & SSIM $\uparrow$ \\
\hline
\noalign{\smallskip}
w/o non-linear $\Delta_v$ & 31.86 & 0.974 & 32.19 & 0.975 & 30.56 & 0.965 \\
w/o AO (pose-dep color)  & 32.94 & 0.980 & 33.20 & 0.980 & 30.57 & 0.968 \\
w/o r.f. interplation & 33.02 & 0.980 & 33.31 & 0.981 & \textbf{30.72} & \textbf{0.969} \\
\hline\noalign{\smallskip}
Ours & \textbf{33.11} & \textbf{0.981} & \textbf{33.35} & \textbf{0.981} & 30.69 & \textbf{0.969} \\
\hline\hline
\end{tabular}
}
\end{center}
\vspace*{-0.8em}
\caption{\textit{Model ablations on the ZJU Mocap subjects}.}
\label{tab:abl}
\vspace*{-5em}
\end{table}
\setlength{\tabcolsep}{1.4pt}

\section{Discussion}
\label{sec:conclusion}

In this paper, we proposed a volumetric representation for articulated actors based on learned skinning, shape, and appearance.
We also model pose-dependent deformation and shading effects.
Extensive evaluations demonstrate that our approach consistently outperforms previous methods when generalizing to out-of-distribution unseen poses.
Our approach can recover much more accurate dense correspondences across different poses and views than prior works, enabling content editing applications.
Moreover, it does not require any body templates, enabling applications for creatures beyond humans. 
While our approach has clear advantages, there are few limitations.
First, our method trains much slower (5 to 8 times)
than the baselines, due to the nature of the root finding process for inverse deformation.
A future direction could be to use invertible neural networks to avoid the root finding process.
Second, even though our forward LBS ensures generalization to unseen poses, pose-dependent \emph{non-linear} deformation and shading effects are still challenging to estimate correctly for unseen poses.
Such effects are fundamentally challenging to model, particularly for lighting-dependent shading. 
An interesting future direction could be to model these effects across multiple subjects so that information from all subjects can be used to improve non-linear deformation model performance. 

\clearpage
\appendix

\section{Inverse Skinning Gradients.}
As described in Sec.~3.3, for each sample $\mathbf{x}_v$ in the view space, we find its canonical correspondence $\mathbf{x}_c$ through root finding:
\begin{align}
    \text{Find } \mathbf{x}_c^*, \quad
    \text{s.t. } f(\mathbf{x}_c^*) = LBS(\mathbf{w}(\mathbf{x}_c^*; \Theta_s), \mathbf{P}, \mathbf{x}_c^*) + \Delta_w(\mathbf{x}_c^*, \mathbf{P}; \Theta_\Delta) - \mathbf{x}_v = \mathbf{0}
\end{align}
In order to optimize the skinning deformation defined by ($F_{\Theta_s}$, $F_{\Theta_\Delta}$), we need to determine the gradients of the overall loss $\mathcal{L}$ w.r.t the network parameters ($\Theta_s$, $\Theta_\Delta$):
\begin{align}
    \frac{\partial \mathcal{L}}{\partial \Theta_s} = \left[ \frac{\partial \mathcal{L}}{\partial \mathbf{x}^*_c} \right] \left[ \frac{\partial \mathbf{x}^*_c}{\partial \Theta_s} \right].
    \qquad
    \frac{\partial \mathcal{L}}{\partial \Theta_\Delta} = \left[ \frac{\partial \mathcal{L}}{\partial \mathbf{x}^*_c} \right] \left[ \frac{\partial \mathbf{x}^*_c}{\partial \Theta_\Delta} \right].
\end{align}
The first term [$\partial \mathcal{L} / \partial \mathbf{x}^*_c$] can be easily calculated through back-propagation.
The second terms [$\partial \mathbf{x}^*_c / \partial \Theta_s$] and [$\partial \mathbf{x}^*_c / \partial \Theta_\Delta$] can be calculated analytically via implicit differentiation~\cite{chen2021snarf}:

\begin{align}
\text{Let,}\quad & \mathrm{\Gamma}(\mathbf{x}_c^*, \mathbf{P}; \Theta_s, \Theta_\Delta) = LBS(\mathbf{w}(\mathbf{x}_c^*; \Theta_s), \mathbf{P}, \mathbf{x}_c^*) + \Delta_w(\mathbf{x}_c^*, \mathbf{P}) \\
\text{Then,}\quad & \mathrm{\Gamma}(\mathbf{x}_c^*, \mathbf{P}; \Theta_s, \Theta_\Delta) - \mathbf{x}_v = \mathbf{0} \\
\Leftrightarrow\quad & \frac{\partial \mathrm{\Gamma}(\mathbf{x}_c^*, \mathbf{P}; \Theta_s, \Theta_\Delta)}{\partial \Theta_s} + \frac{\partial \mathrm{\Gamma}(\mathbf{x}_c^*, \mathbf{P}; \Theta_s, \Theta_\Delta)}{\partial \mathbf{x}^*_c} \cdot \frac{\partial \mathbf{x}^*_c}{\partial \Theta_s} = \mathbf{0} \\
\Leftrightarrow\quad & \frac{\partial \mathbf{x}^*_c}{\partial \Theta_s} = - \left[ \frac{\partial \mathrm{\Gamma}(\mathbf{x}_c^*, \mathbf{P}; \Theta_s, \Theta_\Delta)}{\partial \mathbf{x}^*_c} \right] ^{-1} \left[ \frac{\partial \mathrm{\Gamma}(\mathbf{x}_c^*, \mathbf{P}; \Theta_s, \Theta_\Delta)}{\partial \Theta_s} \right] \\
\Leftrightarrow\quad & \frac{\partial \mathbf{x}^*_c}{\partial \Theta_s} = - \left[ \frac{\partial \mathbf{x}_v}{\partial \mathbf{x}^*_c} \right] ^{-1} \left[ \frac{\partial \mathbf{x}_v}{\partial \Theta_s} \right]  \\
\text{Similarly,}\quad & \frac{\partial \mathbf{x}^*_c}{\partial \Theta_\Delta} = - \left[ \frac{\partial \mathbf{x}_v}{\partial \mathbf{x}^*_c} \right] ^{-1} \left[ \frac{\partial \mathbf{x}_v}{\partial \Theta_\Delta} \right]
\end{align}

\section{Dataset Splits and Pose Clustering.}
\begin{wrapfigure}{r}{0.34\textwidth}
\centering
\vspace*{-2.3em}
\includegraphics[width=0.9\linewidth]{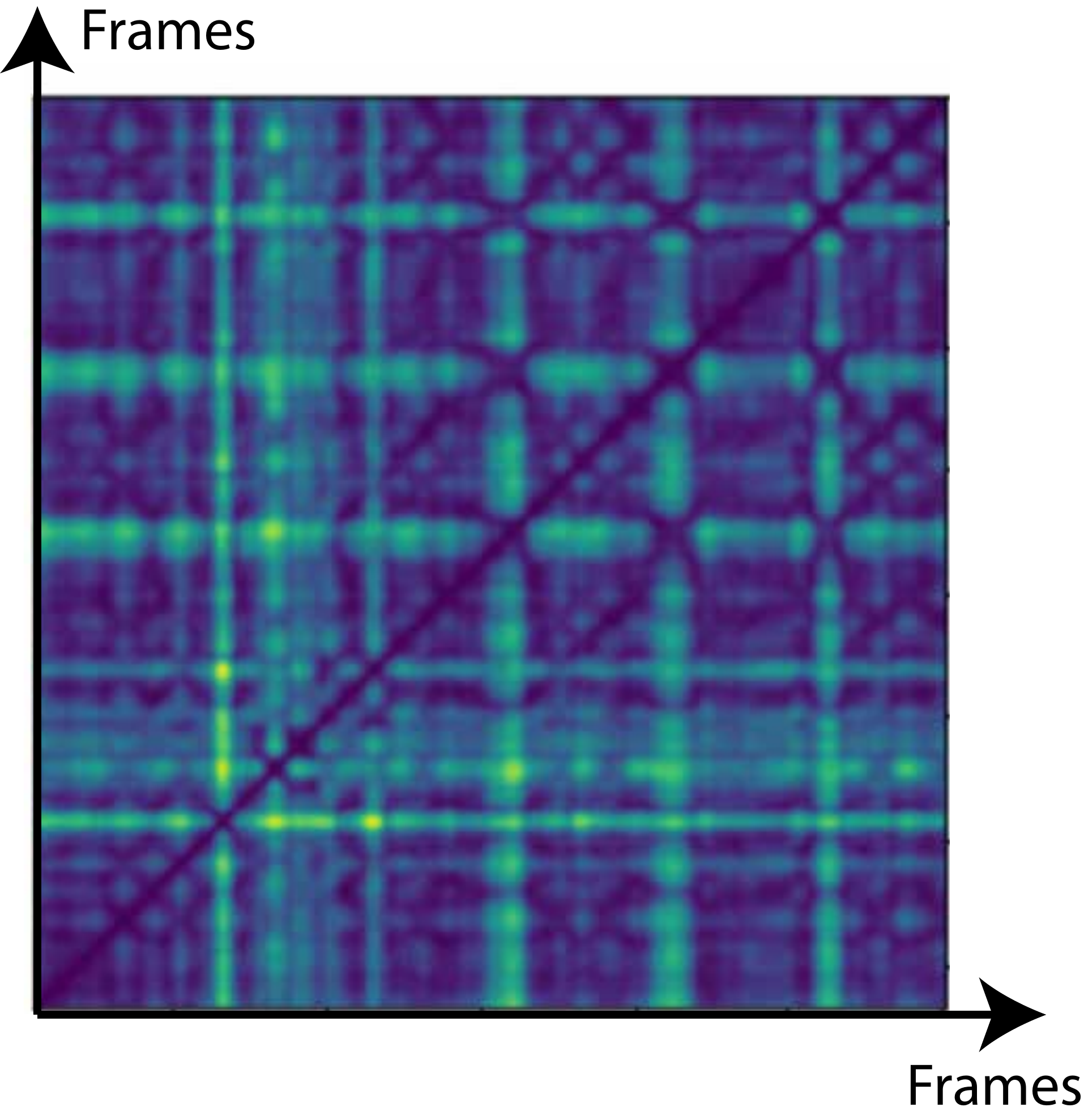}
\vspace*{-1.0em}
\caption{\textit{Pose Similarity Matrix on ZJU Subject 313.}
\vspace*{-2.4em}
}
\label{fig:similarity}
\end{wrapfigure}

As described in Sec~4.1, to avoid similar poses appearing in both the training and the validation set, we split the dataset by clustering the frames based on pose similarity. Fig.~\ref{fig:similarity} shows an example of the pose similarity matrix on ZJU-Mocap subject 313. It clearly shows that this actor moves with a repetitive motion pattern. Thus the previous way~\cite{peng2021animatable,peng2021neural} of splitting the dataset into two chunks with consecutive frames will cover similar poses in both sets, which is not suitable for evaluating the pose generalization ability. This motivated us to introduce our new data split protocol based on pose clustering.

\begin{figure}
\centering
\includegraphics[width=1.0\linewidth]{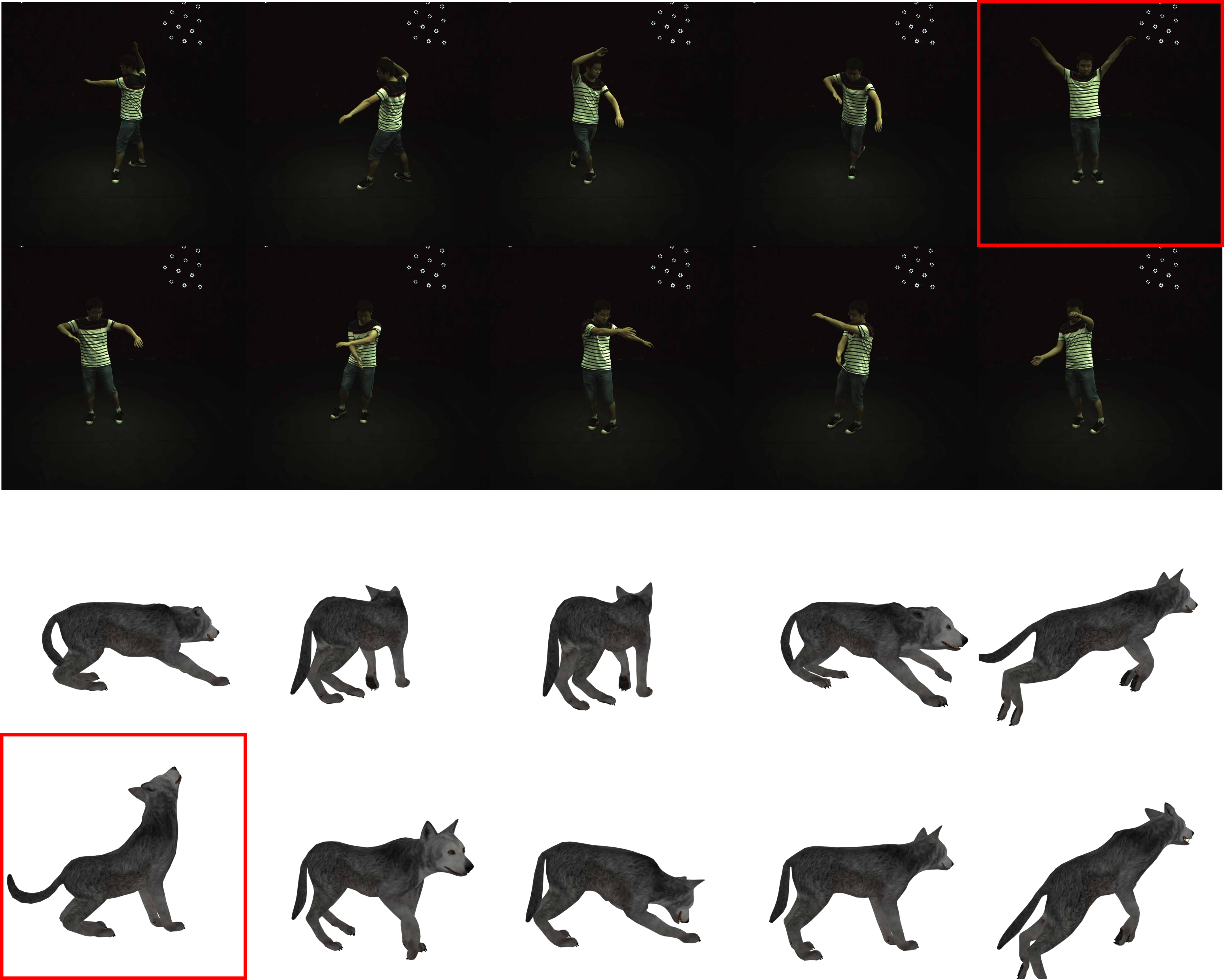}
\caption{\textit{Pose Clustering}. Here we show the K-Medoids clustering results on ZJU-Mocap subject 315 and the Wolf subject. The one cluster marked as red is automatically identified as the most different one thus is selected as the OOD validation set.
}
\label{fig:kmeans}
\end{figure}

Our pose clustering process is as follows: We first disable the global (root) transformation for all poses. Then, the difference of two poses is measured by the Euclidian distance of their corresponding mesh vertices (SMPL mesh for ZJU-Mocap). As we only use pose clustering to construct the dataset splits, the mesh information is considered accessible here. Next the K-Medoids algorithm is adopted to cluster the poses into $K=10$ clusters (Examples shown in Fig.~\ref{fig:kmeans}). Finally we calculate the average difference between the K medoids to find the most different one, which we regard as the out-of-distribution poses to form the OOD validation set.

\section{Implementation Details.}
TAVA employs 
four MLPs ($F_{\Theta_r}$, $F_{\Theta_\Delta}$, $F_{\Theta_r}$, $F_{\Theta_a}$) in our method.
Both, $F_{\Theta_r}$ and $F_{\Theta_\Delta}$, consist of 4 layers with 128 hidden units, with
4-degree positional encoding~\cite{mildenhall2020nerf} on the input coordinates.
$F_{\Theta_r}$ is an 8-layer MLP with 256 hidden units and uses 10-degree integrated positional encoding, similar to Mip-NeRF~\cite{barron2021mip}. $F_{\Theta_a}$ is a single-layer MLP with  128 hidden units that connects to the $8$-th layer of $F_{\Theta_r}$.
We follow the hyper-parameters in Mip-NeRF~\cite{barron2021mip} for the volume rendering, where $64$ samples are drawn for each ray at both coarse and fine level. 

\section{Baseline Implementation Details.}

As described in Sec.~4.2, for the template-based baselines Animatable-NeRF~\cite{peng2021animatable} and NeuralBody~\cite{peng2021neural}, we use their official implementations. For the template-free baselines NARF~\cite{noguchi2021neural} and A-NeRF~\cite{su2021nerf}, we re-implemented them in our code base for fair comparison. We also carefully adapt their official implementations to the ZJU-Mocap dataset to verify our re-implementation. As shown in Tab.~\ref{tab:reimpl}, our re-implementation achieves better performance than the official implementations on ZJU-Mocap subject 313.

\setlength{\tabcolsep}{4pt}
\begin{table}
\begin{center}
\resizebox{0.9\linewidth}{!}{
\begin{tabular}{l cc cc cc}
\hline\hline\noalign{\smallskip}
& \multicolumn{2}{c}{Novel-view} & \multicolumn{2}{c}{Novel-pose (ind)} & \multicolumn{2}{c}{Novel-pose (ood)} \\
& PSNR $\uparrow$ & SSIM $\uparrow$ & PSNR $\uparrow$ & SSIM $\uparrow$ & PSNR $\uparrow$ & SSIM $\uparrow$ \\
\hline
\noalign{\smallskip}
A-NeRF(official) & 28.36 & 0.945 & 28.80 & 0.947 & 27.69 & 0.928 \\
A-NeRF(re-impl.) & \textbf{32.14} & \textbf{0.976} & \textbf{33.39} & \textbf{0.980} &\textbf{28.53} & \textbf{0.953} \\
NARF(official) & 30.65 & 0.962 & 32.22 & 0.969 & \textbf{28.10} & 0.944 \\
NARF(re-impl.) & \textbf{33.17} & \textbf{0.979} & \textbf{34.67} & \textbf{0.983} & 28.05 & \textbf{0.951} \\
\hline\hline
\end{tabular}
}
\end{center}
\caption{\textit{Re-implementation verification on ZJU-Mocap Subject 313}.}
\label{tab:reimpl}
\vspace*{-2em}
\end{table}
\setlength{\tabcolsep}{1.4pt}

\section{Visualizations for Skinning Weights.}

\begin{figure}[h!]
\centering
\includegraphics[width=1.0\linewidth]{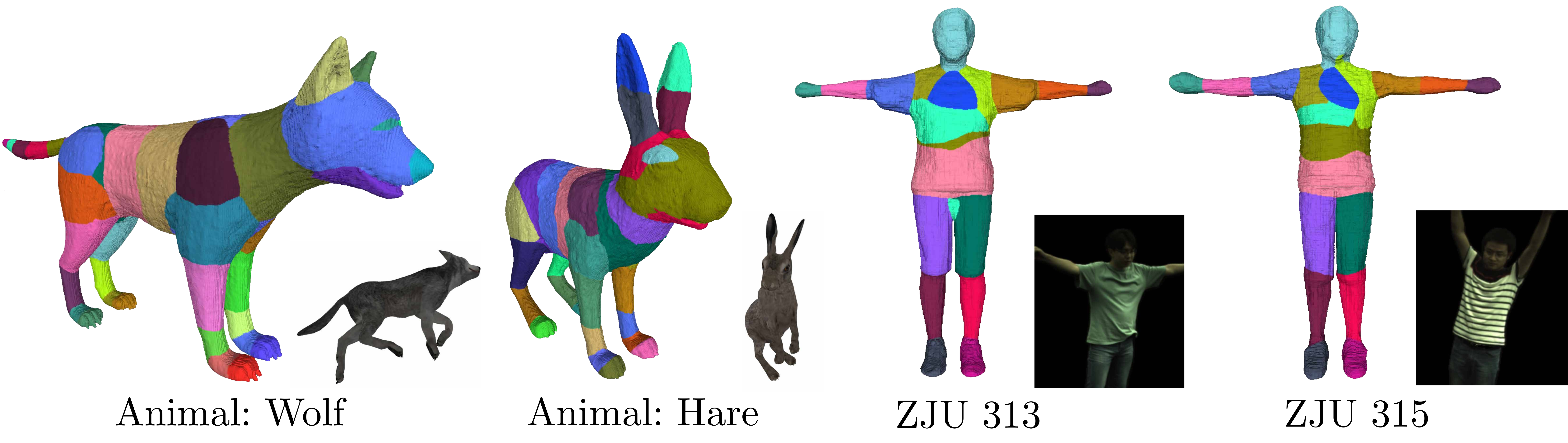}
\caption{\textit{Learned skinning weights and canonical geometry.} Color denotes the top-1 bone from skinning weights.}
\label{fig:skinning_weights}
\end{figure}

Fig.~\ref{fig:skinning_weights} shows results of our learned skinning weights and canonical geometry for animal subjects and ZJU-Mocap data. The surfaces are extracted with marching cube algorithm with threshold $5.0$ on the density field.

\section{Challenges in ZJU-Mocap Dataset.}

\begin{figure}
\centering
\includegraphics[width=0.9\linewidth]{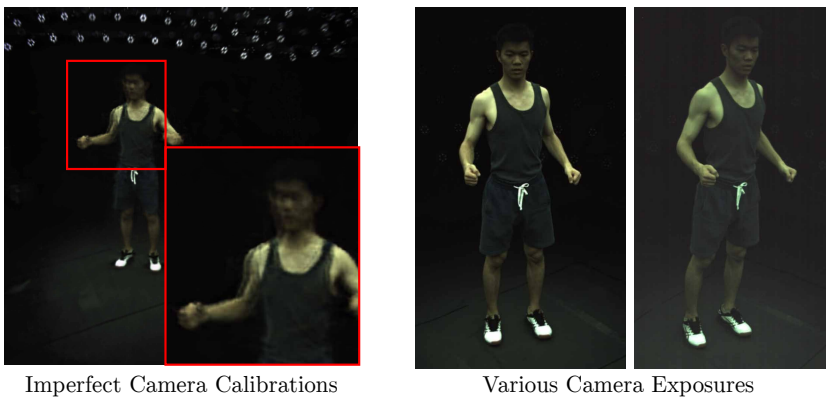}
\caption{\textit{Challenges in ZJU-Mocap Dataset}. \textbf{Left}: We here train a standard NeRF~\cite{yu2021plenoxels} on a single frame with all the views. Imperfect camera calibrations cause the ghost effects on some views. \textbf{Right}: We here compare two groundtruth images side by side for the same subject with different cameras.
}
\label{fig:issues}
\end{figure}

The ZJU-Mocap dataset has become an increasingly popular dataset to study human performance capture, reconstruction, and neural rendeirng~\cite{peng2021animatable,peng2021neural,weng2022humannerf}. Yet we notice that there are a few issues in this dataset that are neither addressed nor mentioned in the previous works, including \emph{imperfect camera calibrations} and \emph{various camera exposures} (as shown in Fig.~\ref{fig:issues}). \textbf{We also get acknowledged from the authors of ZJU-Mocap Dataset on those issues.} We discovered these issues after the submission so they are not considered in our designs. Yet they greatly affect both our performance and the baselines'. We believe it is worth to point them out so that they can be considered in the future research. 

\section{Per-subject Breakdown Comparisons.}

We also report a per-subject breakdown of the quantitative metrics against all baseline methods in Tab.~\ref{tab:breakdown_zju} and Tab.~\ref{tab:breakdown_anim}.

\setlength{\tabcolsep}{4pt}
\begin{table}
\begin{center}
\resizebox{1.0\linewidth}{!}{
\begin{tabular}{l cc cc cc}
\toprule
& \multicolumn{2}{c}{Novel-view} & \multicolumn{2}{c}{Novel-pose (ind)} & \multicolumn{2}{c}{Novel-pose (ood)} \\
& PSNR $\uparrow$ & SSIM $\uparrow$ & PSNR $\uparrow$ & SSIM $\uparrow$ & PSNR $\uparrow$ & SSIM $\uparrow$ \\
\midrule
\noalign{\smallskip}
\emph{\underline{\textbf{Subject 313}}} & & & & & & \\
Animatable-NeRF~\cite{peng2021animatable} & 29.15 & 0.967 & 28.67 & 0.968 & 28.00 & 0.952 \\
NeuralBody~\cite{peng2021neural} & 33.81 & 0.982 & 34.20 & 0.984 & 28.33 & 0.958 \\
Pose-NeRF & 31.94 & 0.974 & 33.22 & 0.979 & 27.03 & 0.941 \\
A-NeRF~\cite{su2021nerf}  & 32.14 & 0.976 & 33.39 & 0.980 & 28.53 & 0.953\\
NARF~\cite{noguchi2021neural}  & 33.17 & 0.979 & 34.67 & 0.983 & 28.05 & 0.951 \\ 
Ours & 33.14 & 0.979 & 34.51 & 0.984 & 29.28 & 0.957 \\
\\
\emph{\underline{\textbf{Subject 315}}} & & & & & & \\
Animatable-NeRF~\cite{peng2021animatable} & 27.62 & 0.962 & 26.50 & 0.956 & 25.52 & 0.949 \\
NeuralBody~\cite{peng2021neural} & 31.41 & 0.982 & 30.35 & 0.984 & 25.87 & 0.957 \\
Pose-NeRF & 28.96 & 0.970 & 28.81 & 0.969 & 24.30 & 0.930 \\
A-NeRF~\cite{su2021nerf}  & 29.67 & 0.974 & 29.46 & 0.973 & 26.82 & 0.959 \\
NARF~\cite{noguchi2021neural}  & 30.18 & 0.977 & 29.96 & 0.976 & 27.05 & 0.960 \\ 
Ours & 30.84 & 0.980 & 30.61 & 0.979 & 26.55 & 0.960 \\
\\
\emph{\underline{\textbf{Subject 377}}} & & & & & & \\
Animatable-NeRF~\cite{peng2021animatable} & 32.17 & 0.979 & 30.20 & 0.974 & 28.95 & 0.969 \\
NeuralBody~\cite{peng2021neural} & 33.86 & 0.985 & 32.96 & 0.983 & 31.55 & 0.978 \\
Pose-NeRF & 32.34 & 0.978 & 32.14 & 0.978 & 29.58 & 0.970 \\
A-NeRF~\cite{su2021nerf}  & 32.62 & 0.980 & 32.53 & 0.980 & 31.77 & 0.978 \\
NARF~\cite{noguchi2021neural}  & 32.87 & 0.982 & 32.83 & 0.981 & 31.83 & 0.978 \\ 
Ours & 33.08 & 0.982 & 33.05 & 0.982 & 32.26 & 0.980 \\
\\
\emph{\underline{\textbf{Subject 386}}} & & & & & & \\
Animatable-NeRF~\cite{peng2021animatable} & 34.07 & 0.975 & 32.00 & 0.967 & 32.23 & 0.974 \\
NeuralBody~\cite{peng2021neural} & 36.55 & 0.985 & 36.19 & 0.984 & 35.57 & 0.983 \\
Pose-NeRF & 34.30 & 0.977 & 34.19 & 0.977 & 32.81 & 0.973 \\
A-NeRF~\cite{su2021nerf}  & 35.37 & 0.981 & 35.23 & 0.980 & 34.50 & 0.979 \\
NARF~\cite{noguchi2021neural}  & 35.53 & 0.981 & 35.39 & 0.981 & 35.46 & 0.982 \\ 
Ours & 35.38 & 0.981 & 35.25 & 0.980 & 34.68 & 0.980 \\

\bottomrule
\end{tabular}
}
\end{center}
\caption{\textit{Per-Subject Comparisons on the ZJU Mocap Dataset}.}
\label{tab:breakdown_zju}
\end{table}
\setlength{\tabcolsep}{1.4pt}

\setlength{\tabcolsep}{4pt}
\begin{table}
\begin{center}
\resizebox{1.0\linewidth}{!}{
\begin{tabular}{l cc ccc ccc}
\toprule
& \multicolumn{2}{c}{Novel-view} & \multicolumn{3}{c}{Novel-pose (ind)} & \multicolumn{3}{c}{Novel-pose (ood)} \\
& PSNR $\uparrow$ & SSIM $\uparrow$  & PSNR $\uparrow$ & SSIM $\uparrow$ & P2P $\downarrow$ & PSNR $\uparrow$ & SSIM $\uparrow$ & P2P $\downarrow$ \\
\midrule
\noalign{\smallskip}

\emph{\underline{\textbf{Subject Hare}}} & & & & & & & & \\
Pose-NeRF & 23.97 & 0.949 & 22.28 & 0.942 & 197.01 & 15.35 & 0.925 & 100.72 \\
A-NeRF~\cite{su2021nerf} & 31.33 & 0.974 & 31.28 & 0.974 & 35.44 & 23.00 & 0.960 & 26.53 \\
NARF~\cite{noguchi2021neural} & 36.45 & 0.986 & 36.56 & 0.986 & 10.90 & 29.40 & 0.979 & 5.64 \\
Ours & 37.35 & 0.990 & 37.57 & 0.990 & 5.04 & 35.24 & 0.989 & 3.91 \\
\\
\emph{\underline{\textbf{Subject Wolf}}} & & & & & & & & \\
Pose-NeRF & 22.83 & 0.946 & 21.57 & 0.941 & 197.21 & 17.90 & 0.925 & 76.97 \\
A-NeRF~\cite{su2021nerf} & 31.20 & 0.979 & 31.16 & 0.979 & 27.60 & 28.31 & 0.974 & 11.55 \\
NARF~\cite{noguchi2021neural} & 36.64 & 0.989 & 36.74 & 0.990 & 7.65 & 32.43 & 0.985 & 11.27 \\
Ours & 37.26 & 0.992 & 37.33 & 0.992 & 3.57 & 36.30 & 0.992 & 2.85 \\

\bottomrule
\end{tabular}
}
\end{center}
\caption{\textit{Per-Subject Comparisons on the animal subje\textbf{}cts Hare and Wolf}.}
\label{tab:breakdown_anim}
\end{table}
\setlength{\tabcolsep}{1.4pt}

\clearpage

%
%
\bibliographystyle{splncs04}
\bibliography{egbib}
\end{document}